\documentclass[letterpaper, 10 pt, conference]{ieeeconf}  
\usepackage{graphicx}

\IEEEoverridecommandlockouts                              

\title{\LARGE \bf
Sthymuli: a Static Educational Robot.\\Leveraging the Thymio II Platform.
}

\author{Manuel Bernal-Lecina$^{1}$, Alejandrina Hernández$^{2}$, Adrien Pannatier$^{1}$, Léa Pereyre$^{1}$ and Francesco Mondada$^{1}$
\thanks{$^{1}$MOBOTS Lab, EPFL,
        Lausanne, Switzerland \newline
        {\tt\small manuel.bernallecina@epfl.ch, adrien.pannatier@epfl.ch, lea.pereyre@epfl.ch, francesco.mondada@epfl.ch}}%
\thanks{$^{2}$ECAL, Lausanne, Switzerland \newline {\tt\small alejandrina.hernandez@ecal.ch}}%
}

\begin{document}

\maketitle
\thispagestyle{empty}
\pagestyle{empty}

\begin{abstract}
The use of robots in education represents a challenge for teachers and a fixed vision of what robots can do for students. This paper presents the development of Sthymuli, a static educational robot designed to explore new classroom interactions between robots, students and teachers. We propose the use of the Thymio II educational platform as a base, ensuring a robust benchmark for a fair comparison of the commonly available wheeled robots and our exploratory approach with Sthymuli. 
This paper outlines the constraints and requirements for developing such a robot, the current state of development and future work.
\end{abstract}

\section{INTRODUCTION}

The integration of robotics in education has shown significant potential to increase student engagement and improve learning outcomes \cite{lopez2021robotics}. Traditional mobile robots, such as the Thymio II, have been effective in teaching programming and STEM concepts \cite{mondada2017physical}. However, the dynamic nature of these robots can present challenges, including maintenance issues, control difficulties and potential distractions in the classroom \cite{mubin2013review}. This scenario adds up to the presence of barriers for technology integration in educational settings \cite{ertmer1999addressing}. Lack of knowledge and prior beliefs make the integration of educational robots a major challenge for teachers. For students, the use of robots that always have the same aesthetics and characteristics leads to a fixed representation of what robots are and therefore how they are perceived. For example, by different genders.  To address these issues, we propose to develop a static robot inspired by robots like Tega and Jibo \cite{westlund2016tega, rane2014study}, two static social robots designed to engage users through natural, intuitive interactions.  

Our primary goal is to create a stable, predictable and easily manageable robotic presence in the classroom. This static robot must facilitate meaningful interactions between students and the robot. For example, by focusing on the aesthetics of the robot. In this way, we hope to address some of the barriers teachers face and help them feel more confident about using robots with their students. 

To achieve these goals, and to measure the results by comparing our static robot with a more traditional wheeled robot, we decided to base our development on the Thymio II platform \cite{mondada2017physical}. This approach will ensure a stable benchmark for the new robot, as all the hardware will be retained, as well as the graphical programming interface. This solution will ensure the quality of the results by a fair comparison of both robotic platforms.

\section{Methodology}


\subsection{Robot requirements for this study}

In order to evaluate the performance of the new robot in educational environments with both teachers (during professional development sessions) and students (in the classroom), the new robot platform must successfully meet the following requirements:

\begin{itemize}
    \item The electronic components of the new robot must be taken from a Thymio II, no additional components should be added to the robot. 
    \item The new robot must be static, as this is the configuration of robots we want to evaluate in the future. 
    \item The robot has to be composed of a base (static) and different modules where the actions of the robot will take place. Having different modules with different functionalities and aesthetics depending on the educational objectives the robot is used for.
    \item The programming interface for the Thymio II and the one for the new robot has to be consistent in terms of typology and usability.
    \item In high-level terms, the new robot has to be able to use the electronic components of the Thymio II platform in interesting and innovative ways while keeping the clean and simple aesthetic line of the Thymio II.
\end{itemize}

\subsection{Design and Development}

The design constraints of the new robot require a design-led approach where iteration plays a central role, helping to integrate new ideas and add improvements with each iteration. We aim to replace the robot's movements with other fascinating aspects based on aesthetics, playfulness and creativity.

For the development, 3D printing technologies play an important role in this design-based approach. They will also facilitate the initial production of prototypes for the testing phases. 

\subsection{Interface}

Thymio II has a variety of programming tools (event-based, block-based and text-based), all of which are embedded in the Thymio Suite. The simplest, and the one most used by primary school children, is called VPL3, an event-based graphical interface that allows actions to be performed when programmed events occur. We decided to reproduce this interface for two reasons. Firstly, our intention is to train primary school teachers, who are the ones least trained in STEM or CS subjects. Secondly, this interface was the easiest to adapt for this purpose.

\section{RESULTS: THE STHYMULI ROBOT} 

\subsection{Prototype ONE}

\begin{figure}[htbp]
    \centerline{\includegraphics[width=0.9\columnwidth]{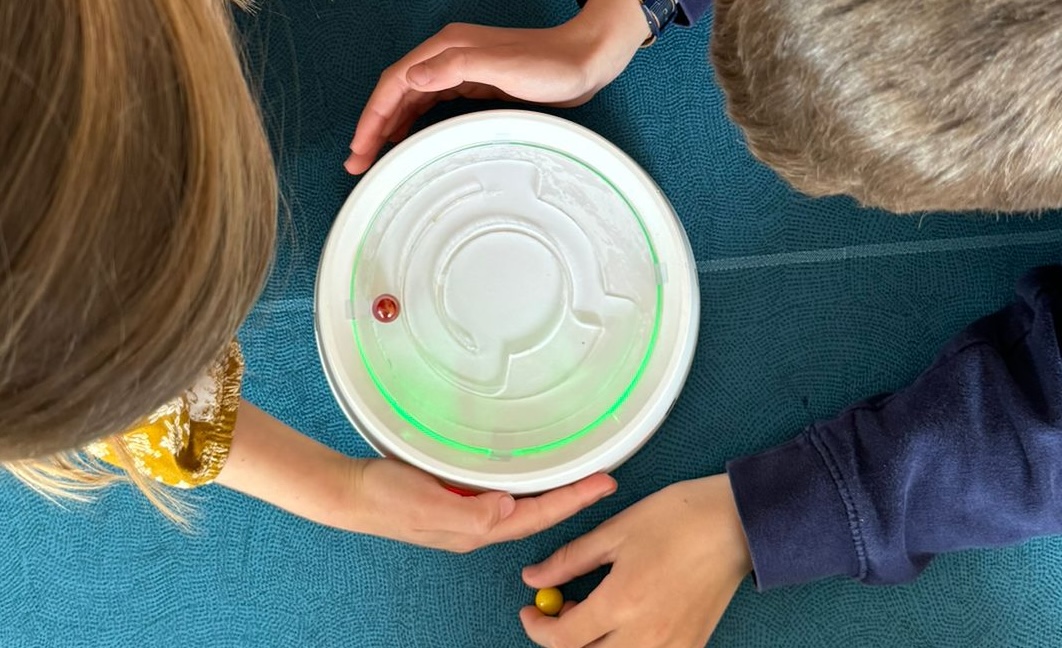}}
    \caption{Testing with kids the interactivity of the first prototype.}
    \label{fig:protoONE}
\end{figure}

Prototype One was designed with a strong focus on validating potential modules around a circular base. Several ideas for modules emerged, but we kept only two under development. One was a biaxial vibration plate, capable of moving small objects in two different axes, allowing a wide range of movements and effects. The second was a flexible spinal arm mounted on a rotating base. Both modules required 2 motors to produce the desired behaviours.
In \ref{fig:protoONE} we show the final prototype with the vibrating plate, including add-ons to increase the potential of this module.

\subsubsection*{Testing round}

As part of the design process we took Prototype ONE to a test environment with three kids of 3, 7 and 10 years of age. The kids interacted with the robot and we observed the placement of their hands as well as the interaction with the module during different modes of use. Figure \ref{fig:protoONE} represents this testing environment.

\subsection{Prototype TWO}

\begin{figure}[h]
    \centerline{\includegraphics[width=0.9\columnwidth]{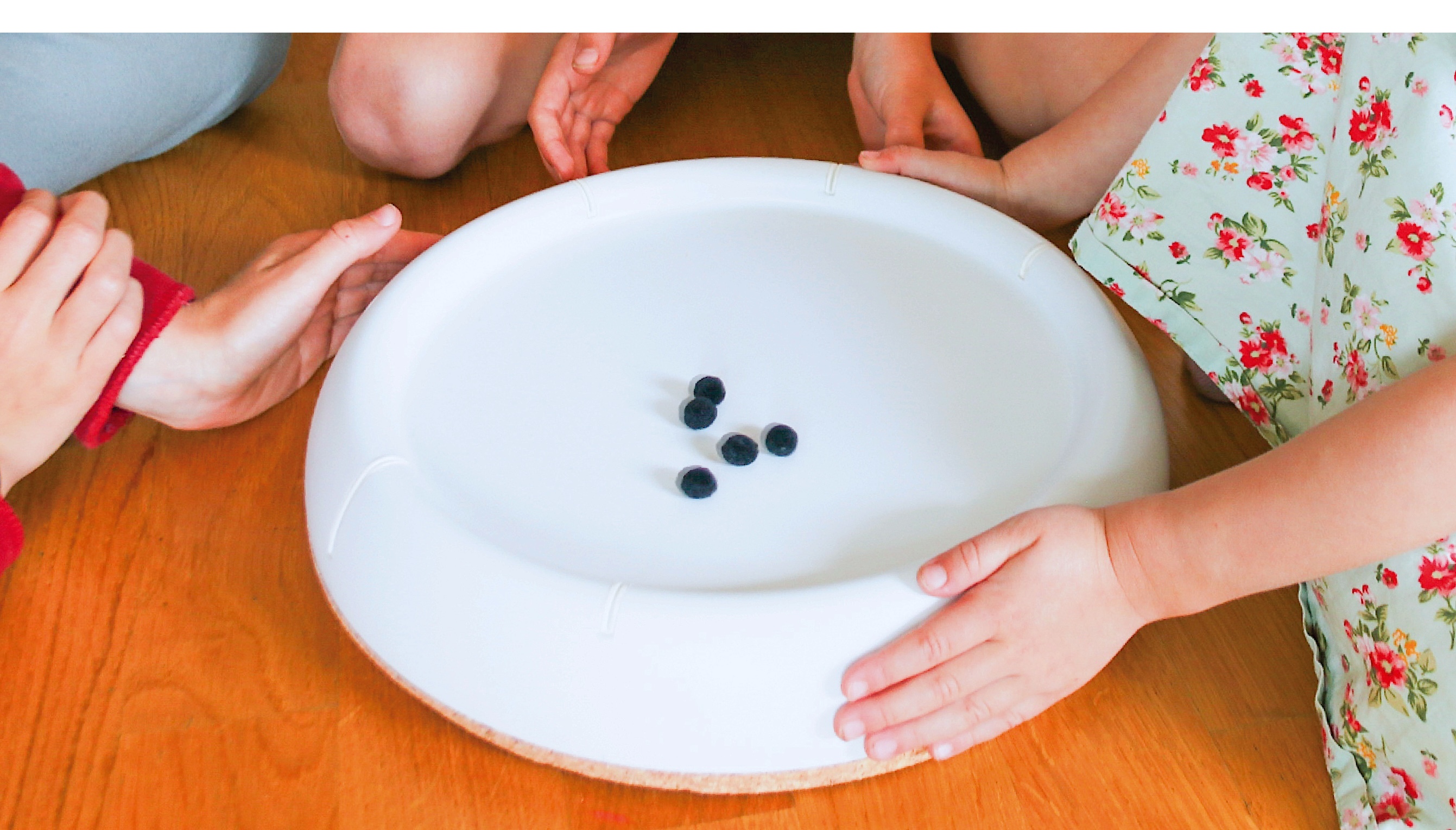}}
    \caption{Second prototype during second testing session.}
    \label{fig:protoTWO}
\end{figure}

Based on the observations  of the first testing round, Prototype TWO focuses on the development of the vibration plate module. The exterior case was enlarged to allow for more kids to interact simultaneously with the robot. The exterior design was smoothed and a cork base was used to isolate the robot from the table or floor. The cork plays also an important role in avoiding the robot to displace on the hard surface bellow it due to the vibrations. Figure \ref{fig:protoTWO} shows the final Sthymuli prototype. 

\subsection{Programming Interface}

\begin{figure}[h]
    \centerline{\includegraphics[width=0.9\columnwidth]{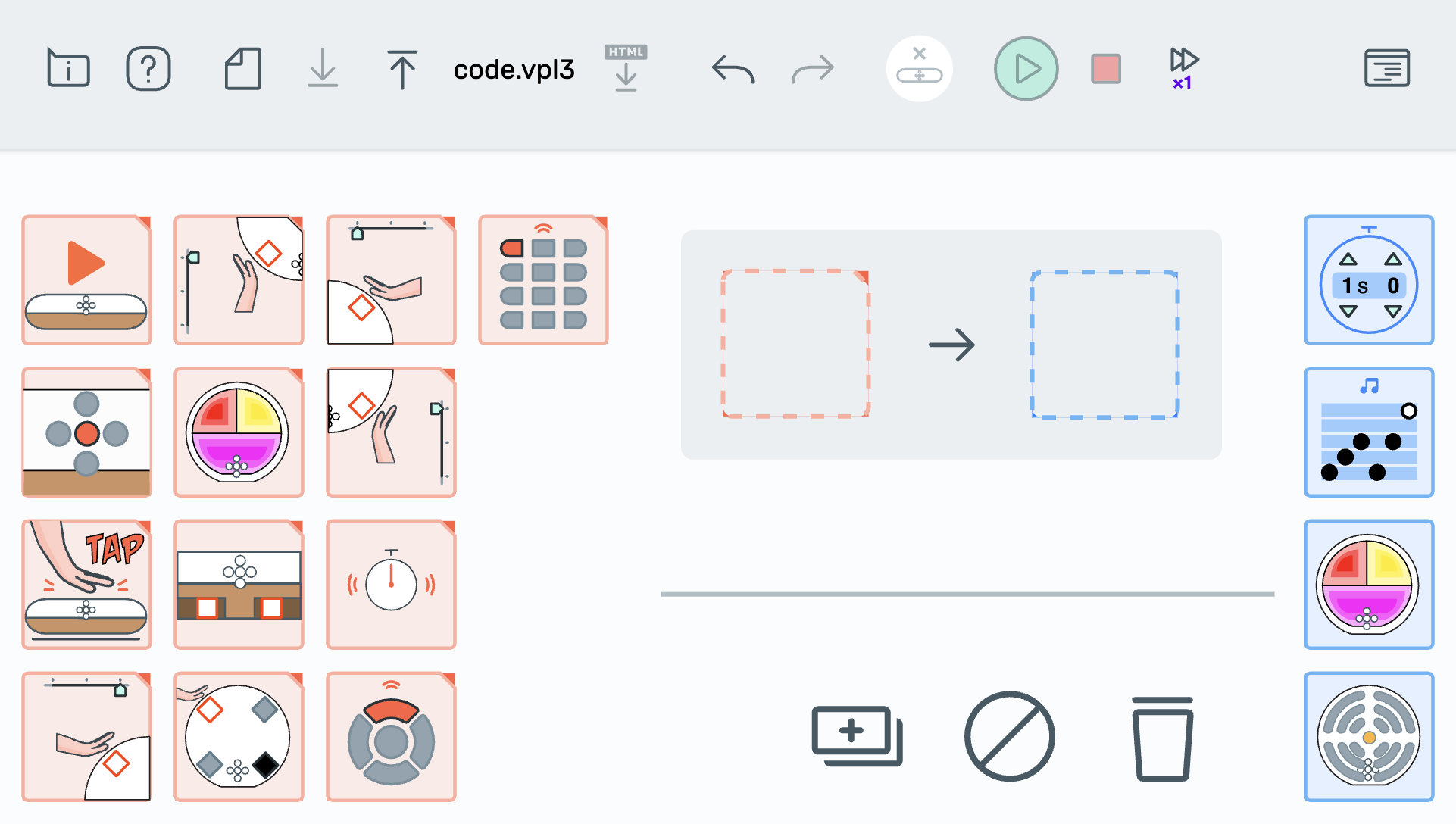}}
    \caption{Final iteration of Sthymuli's VPL3 interface.}
    \label{fig:GUI}
\end{figure}

The Sthymuli VPL2 interface required an adaptation of the icons since the exterior design of Sthymuli was not compatible with the Thymio icons. An adaptation of the programming was also required since Sthymuli does not displace in the environment, new icons for the vibration plate were added. After all, the interface was keep identical as the original VPL3 in terms of usability and connectivity with the robots. In Figure \ref{fig:GUI} we show a close up view of the final interface.

\section{CONCLUSIONS and FUTURE WORK}

We have developed the Sthymuli robot within the constraints required to test its effectiveness in teaching students and training teachers at primary school level. Future work will involve a more elaborate testing phase and a move into real-world scenarios where we will use Thymio II and Sthymuli to teach the same concepts to similar populations. Another line of research will lead us to explore the potential of combining aesthetics and educational robotics to enhance both the teaching and learning experience.

\addtolength{\textheight}{-12cm}   


\bibliographystyle{IEEEtran} 
\bibliography{root}

\end{document}